\documentclass[letterpaper, 10 pt, conference]{ieeeconf}
\IEEEoverridecommandlockouts    %
\usepackage[T1]{fontenc}
\usepackage{newtxtext,newtxmath}

\usepackage{graphics}
\usepackage{graphicx}
\usepackage{amsmath}
\usepackage{booktabs}
\usepackage{color}

\let\labelindent\relax
\usepackage[inline]{enumitem}

\definecolor{tabblue}{HTML}{1F77B4}

\usepackage{hyperref}
\usepackage{cite}
\usepackage{xspace}
\usepackage{multirow}
\usepackage{makecell}

\usepackage[capitalise]{cleveref}
\usepackage{algorithm} 
\usepackage{algpseudocode}
\usepackage{svg}

\usepackage{xcolor}
\definecolor{darkgreen}{RGB}{0,100,0}
\definecolor{forestgreen}{RGB}{34,139,34}
\usepackage{upgreek}

\usepackage{microtype}

\crefname{figure}{fig.}{figs.}
\Crefname{figure}{Fig.}{Figs.}
\crefname{equation}{eq.}{eqs.}
\Crefname{equation}{Eq.}{Eqs.}

\title{\LARGE \bf DisPlace: Discriminative Place Projections for \\ Multi-Reference Visual Place Recognition}

\author{Dhyey Manish Rajani \qquad Michael Milford \qquad Tobias Fischer
\thanks{All authors are with the QUT Centre for Robotics, School of Electrical Engineering and Robotics at the Queensland University of Technology.}
\thanks{This research was partially supported by the QUT Centre for Robotics, ARC Laureate Fellowship FL210100156 to MM, and ARC DECRA Fellowship DE240100149 to TF.}
}

\begin{document}
\thispagestyle{empty}
\pagestyle{empty}

\maketitle

\begin{abstract}

A key challenge in Visual Place Recognition (VPR) is matching query images against reference maps captured under diverse environmental conditions and viewpoints.
While multiple reference traversals improve robustness, existing fusion strategies either aggregate references uniformly or rely on heuristic selection, 
without distinguishing descriptor variations that preserve stable place identity from those caused by changing conditions or viewpoints.
In this paper, we propose DisPlace, a multi-reference VPR framework that fuses multiple reference descriptors into a single compact and discriminative place representation. 
DisPlace formulates descriptor fusion as a generalized eigenvalue problem that maximizes between-place separability while suppressing within-place variation across references, rather than preserving overall descriptor variance. 
Unlike existing multi-reference fusion methods, DisPlace exploits variation across reference traversals to identify which linear combinations of descriptor dimensions preserve place identity and which capture condition- or viewpoint-specific variation.
We evaluate DisPlace on Oxford RobotCar, Nordland, Pittsburgh30k, and Google Landmarks v2 across six state-of-the-art VPR descriptors.
DisPlace outperforms seven multi-reference baselines in 49 out of 54 appearance-varying conditions, consistently improves descriptor-level fusion performance under viewpoint and unstructured settings, and requires less storage during inference than all compared fusion methods.
\end{abstract}

\bstctlcite{bstctl:nodash}
\section{Introduction}
\label{sec:introduction}

\begin{figure}[t]
    \centering
    \includegraphics[width=0.99\linewidth]{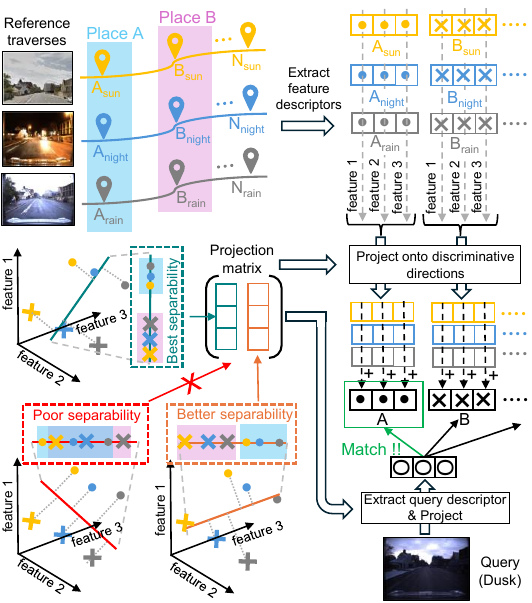}
    \vspace*{-0.2cm}
    \caption{\textbf{Overview of DisPlace for multi-reference VPR.}
    Reference traversals captured under different conditions or viewpoints are encoded into VPR descriptors.
    DisPlace exploits variation across multiple observations of the same place to derive projection directions that preserve stable place-discriminative information while suppressing condition- and viewpoint-induced variation.
    Reference descriptors are projected into this discriminative space and aggregated into a single compact descriptor per place.
    During inference, query descriptors are projected using the same transformation.}
    \label{fig:cover_figure}
    \vspace*{-0.25cm}
\end{figure}

Visual Place Recognition (VPR) enables long-term autonomous navigation by matching a query image against a geo-tagged reference image set of previously visited places~\cite{yin2025generalPRsurvey,schubert2024vprtutorial}.
In real-world deployment, environments undergo substantial appearance variation due to changes in weather, illumination, and season, as well as viewpoint differences across revisits~\cite{schubert2024vprtutorial}.
Consequently, a single reference traversal is unlikely to capture the full range of variations encountered during deployment.

Recent VPR methods have substantially improved robustness through metric learning and discriminative descriptor learning~\cite{berton2022rethinking, berton2023eigenplaces, izquierdo2024optimal, lu2024cricavpr}.
However, these approaches typically operate in a single-reference setting and therefore do not exploit complementary information available across multiple traversals of the same environment.
Maintaining multiple reference traversals, each capturing the environment under different conditions or viewpoints, offers a principled route toward improving recognition performance under changing conditions~\cite{garg2021where, churchill2012practice, molloy2021intelligent}. 
While collecting multiple traversals increases mapping requirements, in many practical deployments such as autonomous vehicle fleets, ride-sharing platforms, and large-scale mapping services, repeated traversals arise naturally as part of normal operation~\cite{hausler2022improveworstcase}, making multi-reference representations both practical and increasingly available at scale.

Existing multi-reference approaches either directly aggregate descriptors using operations such as element-wise summation~\cite{malone2025hyperdimensional}, treating all descriptor dimensions equally, or compare queries against each reference traversal independently and fuse retrieval scores~\cite{fischer2020event, molloy2021intelligent}, requiring all traversals to remain available at inference time.
None of these methods explicitly model how descriptor dimensions vary across references to distinguish stable place information from condition- or viewpoint-specific variation.

In this paper, we address multi-reference VPR by deriving compact per-place representations from multiple reference traversals to improve retrieval robustness while reducing storage and inference complexity. The key challenge is that descriptors of the same place may vary substantially across conditions and viewpoints, while descriptors of different places must remain separable. 

To address this challenge, we propose DisPlace, a multi-reference VPR framework that derives a discriminative projection from variation across reference traversals.
DisPlace formulates descriptor fusion as a generalized eigenvalue problem that maximizes between-place separability while suppressing within-place variation, identifying descriptor directions that preserve stable place information while suppressing condition- and viewpoint-induced variation. Reference descriptors are then projected into this discriminative space and fused into a single compact per-place representation.

The main contribution of this paper is a discriminative fusion framework for multi-reference VPR that explicitly models variation across reference traversals to derive compact and robust place representations. In particular:
\begin{enumerate}
    \item We propose DisPlace, a multi-reference VPR framework that formulates descriptor fusion as a generalized eigenvalue problem to derive discriminative place representations by maximizing between-place separability while suppressing within-place variation across references.
    
    \item We show that DisPlace produces compact fused representations that achieve equivalent or superior retrieval performance across a wide range of descriptor dimensionalities relative to existing multi-reference VPR baselines, while reducing storage and inference complexity.
    
    \item We evaluate DisPlace across four datasets and six state-of-the-art VPR descriptors under appearance, viewpoint, and unstructured variation, demonstrating consistent improvements over existing fusion approaches.
\end{enumerate}
\section{Related Work}
In this section, we review related work on Visual Place Recognition (VPR) and multi-reference approaches for VPR.

\subsection{Visual Place Recognition}

Visual place recognition methods that operate against a single reference traversal have evolved from handcrafted local descriptors such as SIFT~\cite{lowe2004sift} and SURF~\cite{bay2006surf}, aggregated using Bag-of-Words~\cite{csurka2004visual} and VLAD~\cite{jegou2010vlad}, toward deep descriptor learning approaches designed to improve robustness to appearance and viewpoint variation.

Learned global descriptors substantially advanced VPR: NetVLAD~\cite{arandjelovic2016netvlad} combined CNN feature extraction with trainable VLAD aggregation, and follow-up works improved robustness through enhanced training strategies~\cite{berton2022rethinking, berton2023eigenplaces}, feature mixing~\cite{ali2023mixvpr}, optimal transport aggregation~\cite{izquierdo2024optimal}, and large-scale multi-task training~\cite{berton2025megaloc}. However, these approaches all fundamentally operate against a single reference traversal and therefore rely on learned descriptor invariance to handle changing appearance and viewpoint conditions~\cite{yin2025generalPRsurvey, garg2021where, schubert2024vprtutorial}.

Rather than improving single-image descriptors directly, sequence-based methods exploit temporal continuity by filtering or aggregating similarity scores across consecutive frames~\cite{mereu2022learningseq, garg2022seqmatchnet}, while still typically operating against a single reference traversal.
Although sequence matching improves robustness under gradual appearance and viewpoint changes, it requires temporally ordered reference and query data, limiting applicability when sequential information is unavailable~\cite{Vysotska2019effective, Vysotska2016lazy, garg2021where}.

In contrast to approaches that seek robustness from a single reference traversal through learned invariance or temporal filtering, DisPlace exploits complementary observations across multiple reference traversals to explicitly separate stable place-discriminative information from condition- and viewpoint-induced variation within the descriptor space.

\subsection{Multi-Reference Visual Place Recognition}

Enriching the reference map with multiple traversals of the same route, captured under different conditions or viewpoints, provides complementary observations of each place and has been shown to improve recognition robustness over single-reference baselines~\cite{garg2021where}.

A first line of work improves multi-reference VPR by selecting or prioritizing informative reference traversals rather than using all available observations.
Churchill and Newman~\cite{churchill2012practice} introduced the multi-experience paradigm for long-term localization across varying conditions and viewpoints.
Subsequent approaches explored scalable map maintenance~\cite{Linegar2015worksmart, doan2019scalable} and probabilistic reference selection and fusion~\cite{molloy2021intelligent} to improve robustness while controlling computational overhead.

A second line of work instead fuses multiple traversals into a single per-place representation.
Fischer and Milford~\cite{fischer2020event} aggregate similarity scores across reference sets, while Malone~et~al.~\cite{malone2025hyperdimensional} directly fuse descriptors using Hyperdimensional computing to produce a single per-place representation.
Vysotska and Stachniss~\cite{Vysotska2019effective} jointly match query sequences against multiple traversals using a graph-based data association framework.

While existing multi-reference approaches perform reference selection~\cite{Linegar2015worksmart, doan2019scalable, molloy2021intelligent}, sequence-based matching~\cite{Vysotska2019effective, Vysotska2016lazy}, score aggregation~\cite{fischer2020event}, or descriptor fusion~\cite{malone2025hyperdimensional}, they generally treat variation across traversals implicitly through score fusion, traversal selection, or descriptor aggregation.
DisPlace instead explicitly exploits variation across reference traversals to derive discriminative projections that preserve stable place information while suppressing condition- and viewpoint-specific variation, enabling compact and robust multi-reference place representations.
\section{Methodology}

\subsection{Preliminaries}

We consider a setting where a robot has collected $K$ reference traversals of an environment containing $N$ distinct places, with each traversal capturing every place under a different condition or viewpoint.
Each reference image of place $p \in \{1, \ldots, N\}$ under condition $k \in \{1, \ldots, K\}$ yields a $D$-dimensional descriptor $\mathbf{r}_p^k \in \mathbb{R}^D$ extracted using a VPR technique (see~\Cref{subsec:descriptors}).
During inference, a query descriptor $\mathbf{q} \in \mathbb{R}^D$ is extracted and matched against the reference map to identify the most likely place.

\subsection{Proposed Method}

The core challenge in multi-reference VPR is that descriptors of the same place vary substantially across conditions or viewpoints, while descriptors of different places must remain separable regardless of condition or viewpoint.
Naive fusion strategies such as element-wise summation or distance matrix averaging treat all descriptor dimensions equally, irrespective of their discriminative utility.
DisPlace addresses this by deriving a projection $\mathbf{P} \in \mathbb{R}^{D \times n}$ that maximizes between-place separability relative to within-place variation, compressing the descriptor space from $D$ to $n \leq D$ dimensions. 
DisPlace assumes that the variation observed across the reference traversals is broadly representative of the appearance or viewpoint changes encountered at query time. 

\subsubsection{Scatter Matrix Construction}

We first construct two scatter matrices that capture within-place variation across reference conditions and between-place variation across different places.
The mean descriptor for each place is computed as:
\begin{equation}
    \boldsymbol{\mu}_p =
    \frac{1}{K} \sum_{k=1}^{K} \mathbf{r}_p^k
    \in \mathbb{R}^{D},
    \quad p \in \{1, \ldots, N\},
\end{equation}
and the global mean across all places as:
\begin{equation}
    \boldsymbol{\mu} =
    \frac{1}{N} \sum_{p=1}^{N} \boldsymbol{\mu}_p
    \in \mathbb{R}^{D}.
\end{equation}
The within-place scatter matrix
$\mathbf{S_W} \in \mathbb{R}^{D \times D}$
captures descriptor variation across reference conditions:
\begin{equation}
    \mathbf{S_W} =
    \frac{1}{NK}
    \sum_{p=1}^{N}
    \sum_{k=1}^{K}
    (\mathbf{r}_p^k - \boldsymbol{\mu}_p)
    (\mathbf{r}_p^k - \boldsymbol{\mu}_p)^T,
\end{equation}
while the between-place scatter matrix
$\mathbf{S_B} \in \mathbb{R}^{D \times D}$
captures the variation between different places:
\begin{equation}
    \mathbf{S_B} =
    \frac{1}{N}
    \sum_{p=1}^{N}
    (\boldsymbol{\mu}_p - \boldsymbol{\mu})
    (\boldsymbol{\mu}_p - \boldsymbol{\mu})^T.
\end{equation}

\subsubsection{Discriminative Projection}

Given $\mathbf{S_W}$ and $\mathbf{S_B}$, we seek a projection direction $\boldsymbol{\upomega} \in \mathbb{R}^D$ such that descriptors of the same place cluster together while descriptors of different places remain separable after projection.
Projection onto $\boldsymbol{\upomega}$ yields between-place variance $\boldsymbol{\upomega}^T \mathbf{S_B} \boldsymbol{\upomega}$ and within-place variance $\boldsymbol{\upomega}^T \mathbf{S_W} \boldsymbol{\upomega}$. The optimal projection direction therefore maximizes their ratio. 
Since this ratio is invariant to the magnitude of $\boldsymbol{\upomega}$, we impose $\boldsymbol{\upomega}^T \mathbf{S_W} \boldsymbol{\upomega} = 1$ and solve:
\begin{equation}
    \max_{\boldsymbol{\upomega}}
    \;\;
    \boldsymbol{\upomega}^T \mathbf{S_B} \boldsymbol{\upomega}
    \;\;
    \text{subject to}
    \;\;
    \boldsymbol{\upomega}^T \mathbf{S_W} \boldsymbol{\upomega} = 1.
    \label{eq:optimisation_prob}
\end{equation}

We solve~\Cref{eq:optimisation_prob} using the Lagrangian:
\begin{equation}
    \mathcal{L}(\boldsymbol{\upomega}, \lambda)
    =
    \boldsymbol{\upomega}^T \mathbf{S_B} \boldsymbol{\upomega}
    -
    \lambda
    \left(
    \boldsymbol{\upomega}^T \mathbf{S_W} \boldsymbol{\upomega}
    -
    1
    \right),
\end{equation}
where $\lambda$ is a Lagrange multiplier.
Setting the derivative with respect to $\boldsymbol{\upomega}$ to zero yields:
\begin{equation}
    \frac{\partial \mathcal{L}}
    {\partial \boldsymbol{\upomega}}
    =
    2\mathbf{S_B}\boldsymbol{\upomega}
    -
    2\lambda\mathbf{S_W}\boldsymbol{\upomega}
    =
    0,
\end{equation}
which simplifies to the generalized eigenvalue problem:
\begin{equation}
    \mathbf{S_B} \mathbf{v}
    =
    \lambda \mathbf{S_W} \mathbf{v},
    \label{eq:geig}
\end{equation}
where $\lambda_i \in \mathbb{R}$ and $\mathbf{v}_i \in \mathbb{R}^D$ denote the $i$-th eigenvalue and eigenvector, respectively.
Each eigenvalue $\lambda_i$ quantifies the ratio of between-place to within-place variance along its corresponding eigenvector, where larger eigenvalues indicate more discriminative projection directions.

The eigenvectors are sorted in descending order of eigenvalue magnitude, and the projected dimensionality $n$ is selected as the minimum number of components whose cumulative eigenvalue sum captures a fraction $\tau$ of the total discriminative variance:
\begin{equation}
    n =
    \min \left\{
    m
    \,:\,
    \frac{
    \sum_{i=1}^{m} \lambda_i
    }{
    \sum_{\lambda_i > 0} \lambda_i
    }
    \geq
    \tau
    \right\},
    \label{eq:n_select}
\end{equation}
where the denominator sums only over positive eigenvalues.
The top $n$ eigenvectors form the projection matrix:
\begin{equation}
    \mathbf{P}
    =
    [\mathbf{v}_1 \mid \mathbf{v}_2 \mid \cdots \mid \mathbf{v}_n]
    \in \mathbb{R}^{D \times n}.
\end{equation}
Unlike unsupervised dimensionality reduction methods that preserve overall descriptor variance, the derived projection explicitly prioritizes directions that maximize between-place separability relative to within-place variation. 
In this paper, we set $\tau = 0.95$; an ablation study is provided in Section~\ref{subsec:ablation}.

\subsubsection{Reference Descriptor Fusion}

With $\mathbf{P}$ computed, each place $p$ is represented by a single fused descriptor $\mathbf{\hat{h}}_p \in \mathbb{R}^n$.
The $K$ reference descriptors are projected into the discriminative space, summed, and L2-normalized:
\begin{equation}
    \mathbf{\hat{h}}_p
    =
    \frac{
    \sum_{k=1}^{K} \mathbf{r}_p^k \mathbf{P}
    }{
    \left\|
    \sum_{k=1}^{K} \mathbf{r}_p^k \mathbf{P}
    \right\|_2
    }.
    \label{eq:fused}
\end{equation}
Unlike descriptor aggregation methods such as the current state-of-the-art HOPS~\cite{malone2025hyperdimensional}, aggregation in DisPlace occurs in the derived discriminative space rather than the original descriptor space, yielding a more discriminative and compact per-place representation. 

\subsubsection{Query Matching}
\label{method:query_matching}

During inference, the query descriptor $\mathbf{q} \in \mathbb{R}^D$ is projected into the same discriminative space using $\mathbf{P}$ and L2-normalized:
\begin{equation}
    \mathbf{\hat{q}}
    =
    \frac{
    \mathbf{q}\mathbf{P}
    }{
    \|
    \mathbf{q}\mathbf{P}
    \|_2
    }
    \in \mathbb{R}^n.
\end{equation}
The predicted place $p^*$ is then retrieved as the fused descriptor with the highest cosine similarity:
\begin{equation}
    p^*
    =
    \operatorname*{\arg\max}_{p \in \{1,\ldots,N\}}
    \mathbf{\hat{h}}_p \mathbf{\hat{q}}.
    \label{eq:match}
\end{equation}

DisPlace performs fusion offline and stores a single fused descriptor per place during inference.
Consequently, storage and matching complexity scale with $O(N \cdot n)$ rather than $O(K \cdot N \cdot D)$ as in methods that retain all reference traversals.
A detailed analysis of inference-time complexity and storage requirements is provided in Section~\ref{subsec:complexity}.
\section{Experimental Setup}
\label{sec:experimentalsetup}

\setlength\tabcolsep{0.5mm}

\begin{table*}[!ht]
\caption{Recall@1 on Nordland dataset. Rows are grouped into single reference (top) baselines, and score-level (middle) and descriptor-level (bottom) multi-reference baselines. Best single reference result is \underline{underlined}, best multi-reference result is \textbf{bolded}. DisPlace outperforms the best single reference results in 24/24 cases and the best multi-reference baseline in 24/24 cases.}
\vspace*{-0.1cm}
\centering
\scriptsize
\begin{tabular}{l@{\hspace{2mm}}*{4}{c}|*{4}{c}|*{4}{c}|*{4}{c}|*{4}{c}|*{4}{c}}
\toprule
\textbf{Queries $\rightarrow$}
& \rotatebox{90}{\raisebox{0.075cm}{\textbf{Fall}}} & \rotatebox{90}{\raisebox{0.075cm}{\textbf{Spring}}} & \rotatebox{90}{\raisebox{0.075cm}{\textbf{Summer}}} & \rotatebox{90}{\raisebox{0.075cm}{\textbf{Winter}}}
& \rotatebox{90}{\raisebox{0.075cm}{\textbf{Fall}}} & \rotatebox{90}{\raisebox{0.075cm}{\textbf{Spring}}} & \rotatebox{90}{\raisebox{0.075cm}{\textbf{Summer}}} & \rotatebox{90}{\raisebox{0.075cm}{\textbf{Winter}}}
& \rotatebox{90}{\raisebox{0.075cm}{\textbf{Fall}}} & \rotatebox{90}{\raisebox{0.075cm}{\textbf{Spring}}} & \rotatebox{90}{\raisebox{0.075cm}{\textbf{Summer}}} & \rotatebox{90}{\raisebox{0.075cm}{\textbf{Winter}}}
& \rotatebox{90}{\raisebox{0.075cm}{\textbf{Fall}}} & \rotatebox{90}{\raisebox{0.075cm}{\textbf{Spring}}} & \rotatebox{90}{\raisebox{0.075cm}{\textbf{Summer}}} & \rotatebox{90}{\raisebox{0.075cm}{\textbf{Winter}}}
& \rotatebox{90}{\raisebox{0.075cm}{\textbf{Fall}}} & \rotatebox{90}{\raisebox{0.075cm}{\textbf{Spring}}} & \rotatebox{90}{\raisebox{0.075cm}{\textbf{Summer}}} & \rotatebox{90}{\raisebox{0.075cm}{\textbf{Winter}}}
& \rotatebox{90}{\raisebox{0.075cm}{\textbf{Fall}}} & \rotatebox{90}{\raisebox{0.075cm}{\textbf{Spring}}} & \rotatebox{90}{\raisebox{0.075cm}{\textbf{Summer}}} & \rotatebox{90}{\raisebox{0.075cm}{\textbf{Winter}}}\\
\hline
\textbf{References}
& \multicolumn{4}{c|}{\textbf{CosPlace (512D)}}
& \multicolumn{4}{c|}{\textbf{EigenPlaces (512D)}}
& \multicolumn{4}{c|}{\textbf{MixVPR (4096D)}}
& \multicolumn{4}{c|}{\textbf{NetVLAD (4096D)}}
& \multicolumn{4}{c|}{\textbf{SALAD (8448D)}}
& \multicolumn{4}{c}{\textbf{MegaLoc (8448D)}}\\
\hline
Fall
& - & \underline{68.2} & \underline{74.2} & 42.6
& - & \underline{72.7} & \underline{77.0} & 45.9
& - & \underline{77.8} & \underline{78.0} & 53.7
& - & \underline{29.2} & \underline{44.0} & 7.8
& - & \underline{78.9} & \underline{77.9} & 62.5
& - & 82.7 & \underline{79.5} & 72.8 \\
Spring
& 63.0 & - & 55.9 & \underline{56.0}
& 69.4 & - & 61.1 & \underline{53.6}
& 71.6 & - & 63.9 & \underline{71.7}
& 24.0 & - & 20.8 & \underline{11.2}
& \underline{79.7} & - & 73.9 & \underline{78.5}
& \underline{83.5} & - & 75.9 & \underline{87.9} \\
Summer
& \underline{74.7} & 61.7 & - & 40.8
& \underline{77.4} & 66.5 & - & 44.2
& \underline{78.3} & 72.0 & - & 53.1
& \underline{47.5} & 26.8 & - & 7.6
& 78.4 & 73.4 & - & 61.5
& 79.6 & 75.1 & - & 70.0 \\
Winter
& 32.0 & 51.3 & 29.6 & -
& 39.3 & 58.6 & 36.7 & -
& 43.5 & 68.2 & 42.1 & -
& 5.6 & 10.8 & 5.4 & -
& 60.7 & 78.5 & 60.2 & -
& 72.2 & \underline{88.1} & 69.8 & - \\
\hline

\multicolumn{25}{l}{\textbf{\emph{Score-level baselines}}} \\

dMat. Avg.~\cite{fischer2020event}
& 73.5 & 76.6 & 68.5 & 58.8
& 80.3 & 82.0 & 75.7 & 62.8
& 83.5 & 88.0 & 78.9 & 75.6
& 43.1 & 38.9 & 40.5 & 15.2
& 84.1 & 88.2 & 80.7 & 80.1
& 86.3 & 91.4 & 81.8 & 87.9 \\
dMat. Min.~\cite{fischer2020event}
& 77.1 & 73.6 & 74.8 & 57.5
& 79.0 & 78.2 & 77.5 & 55.2
& 80.4 & 83.8 & 78.5 & 73.0
& 48.7 & 35.6 & 45.3 & 12.6
& 79.9 & 84.1 & 78.4 & 78.9
& 80.9 & 88.4 & 79.9 & 87.4 \\
Std. dMat. Min.
& 70.6 & 69.2 & 66.2 & 56.5
& 78.3 & 76.8 & 75.3 & 57.3
& 79.7 & 83.9 & 75.7 & 72.9
& 39.0 & 29.6 & 35.6 & 12.6
& 80.2 & 84.4 & 78.6 & 77.1
& 81.6 & 89.0 & 80.2 & 87.0 \\
 dMat. Med.~\cite{fischer2020event}
& 63.9 & 68.1 & 58.1 & 51.2
& 70.3 & 73.0 & 63.4 & 55.0
& 71.5 & 76.6 & 66.4 & 65.9
& 27.8 & 29.5 & 25.7 & 12.2
& 78.4 & 80.4 & 74.8 & 71.8
& 82.9 & 85.6 & 76.9 & 81.7 \\
 BSF~\cite{molloy2021intelligent}
& 77.3 & 74.3 & 74.9 & 58.7
& 79.2 & 78.9 & 77.6 & 56.6
& 80.7 & 84.6 & 78.7 & 75.6
& 49.5 & 38.0 & 45.9 & 15.1
& 80.0 & 84.8 & 78.6 & 80.1
& 80.9 & 89.1 & 80.0 & 88.2 \\
\hline

\multicolumn{25}{l}{\textbf{\emph{Descriptor-level baselines}}} \\

Pooling
& 77.1 & 73.6 & 74.8 & 57.5
& 79.0 & 78.2 & 77.5 & 55.2
& 80.4 & 83.8 & 78.5 & 73.0
& 48.7 & 35.6 & 45.3 & 12.6
& 79.9 & 84.1 & 78.4 & 78.9
& 80.9 & 88.4 & 79.9 & 87.4 \\

HOPS~\cite{malone2025hyperdimensional}
& 79.9 & 82.9 & 74.2 & 60.4
& 83.4 & 86.3 & 78.8 & 63.5
& 85.6 & 90.8 & 80.8 & 77.6
& 52.6 & 47.2 & 48.9 & 16.0
& 85.4 & 90.4 & 81.8 & 81.5
& 86.8 & 92.8 & 82.4 & 89.4 \\

DisPlace (Ours)
& \textbf{84.2} & \textbf{87.1} & \textbf{78.7} & \textbf{73.5}
& \textbf{85.4} & \textbf{89.1} & \textbf{80.7} & \textbf{75.6}
& \textbf{87.0} & \textbf{91.7} & \textbf{82.2} & \textbf{83.9}
& \textbf{68.3} & \textbf{62.3} & \textbf{65.3} & \textbf{27.2}
& \textbf{86.2} & \textbf{91.6} & \textbf{82.1} & \textbf{85.5}
& \textbf{87.0} & \textbf{93.2} & \textbf{82.5} & \textbf{90.0} \\

\bottomrule
\end{tabular}
\label{tab:nordland_table}
\vspace*{-0.2cm}
\end{table*}

\subsection{Datasets}
\label{subsec:datasets}

We evaluate DisPlace on four VPR benchmarks spanning appearance, seasonal, viewpoint, and illumination variation across structured and unstructured environments.

\subsubsection{\texorpdfstring{Oxford RobotCar~\cite{maddern20171}}{Oxford RobotCar}} contains over 100 traversals of a fixed route through Oxford captured under varying weather, illumination, and traffic conditions.
Following~\cite{malone2025hyperdimensional, somayeh2025improving, molloy2021intelligent}, we use six traversals corresponding to sun, dusk, rain, night, and two overcast conditions, subsampled at approximately 1\,m intervals, yielding 3,876 images per traversal with direct cross-traversal correspondence.

\subsubsection{\texorpdfstring{Nordland~\cite{sunderhauf2013we}}{Nordland}} records a 729\,km train journey across multiple seasons in Norway and is widely used for evaluating extreme appearance variation.
We use the fall, spring, summer, and winter traversals, subsampled at approximately 2.5\,m intervals, yielding 27,592 images per traversal with direct cross-season correspondence after removing stationary and tunnel sequences, following~\cite{hausler2021patchnetvlad, berton2022benchmark}.

\subsubsection{\texorpdfstring{Pittsburgh30k~\cite{arandjelovic2016netvlad}}{Pittsburgh30k}} is derived from Google Street View imagery and evaluates viewpoint robustness in urban environments.
Following~\cite{arandjelovic2016netvlad, hausler2021patchnetvlad}, we use the test split which contains 417 places, each captured from 24 viewpoints (2 pitch and 12 yaw angles), yielding approximately 10,000 reference images and 6,816 queries.

\subsubsection{\texorpdfstring{Google Landmarks v2 Micro (GLDv2-micro)~\cite{weyand2020gldv2,malone2025hyperdimensional}}{Google Landmarks v2 Micro}} was introduced in~\cite{malone2025hyperdimensional} and comprises 23,294 reference images and 3,103 queries across 3,103 landmark classes, with 7--9 reference images per landmark.
Unlike the traversals in Oxford RobotCar and Nordland, GLDv2-micro contains irregular viewpoint \textit{and} appearance variation from unconstrained real-world images.

\subsection{VPR Descriptors}
\label{subsec:descriptors}

We evaluate DisPlace using six state-of-the-art VPR techniques spanning diverse descriptor architectures and aggregation strategies: CosPlace~\cite{berton2022rethinking}, EigenPlaces~\cite{berton2023eigenplaces}, MegaLoc~\cite{berton2025megaloc}, SALAD~\cite{izquierdo2024optimal}, MixVPR~\cite{ali2023mixvpr}, and NetVLAD~\cite{arandjelovic2016netvlad}. %
All methods use cosine similarity for image matching.

\subsection{Baselines}
\label{subsec:baselines}

We compare against existing multi-reference descriptor-level and score-level VPR approaches.

\textit{Descriptor-level methods} operate directly on reference descriptors.
Pooling stacks all $K$ traversals into a database of $K \times N$ images and performs nearest-neighbor retrieval.
HOPS~\cite{malone2025hyperdimensional} fuses descriptors via Hyperdimensional Computing bundling using element-wise summation to produce a single descriptor per place.

{
\setlength\tabcolsep{0.5mm}

\begin{table*}[t]
\caption{Recall@1 on Oxford RobotCar dataset. 
Rows are grouped into single reference (top) baselines, and score-level (middle) and descriptor-level (bottom) multi-reference baselines. 
Best single reference result is \underline{underlined}, best multi-reference result is \textbf{bolded}. 
DisPlace outperforms the best single reference results in 30/30 cases and the best multi-reference results in 25/30 cases.}
\centering
\scriptsize
\begin{tabular}{l@{\hspace{2mm}}*{5}{c}|*{5}{c}|*{5}{c}|*{5}{c}|*{5}{c}|*{5}{c}}
\toprule
 \textbf{Queries $\rightarrow$}
& \rotatebox{90}{\raisebox{0.075cm}{\textbf{Dusk}}} & \rotatebox{90}{\raisebox{0.075cm}{\textbf{Night}}} & \rotatebox{90}{\raisebox{0.075cm}{\textbf{Overcast}}} & \rotatebox{90}{\raisebox{0.075cm}{\textbf{Overcast2}}} & \rotatebox{90}{\raisebox{0.075cm}{\textbf{Rain}}}
& \rotatebox{90}{\raisebox{0.075cm}{\textbf{Dusk}}} & \rotatebox{90}{\raisebox{0.075cm}{\textbf{Night}}} & \rotatebox{90}{\raisebox{0.075cm}{\textbf{Overcast}}} & \rotatebox{90}{\raisebox{0.075cm}{\textbf{Overcast2}}} & \rotatebox{90}{\raisebox{0.075cm}{\textbf{Rain}}}
& \rotatebox{90}{\raisebox{0.075cm}{\textbf{Dusk}}} & \rotatebox{90}{\raisebox{0.075cm}{\textbf{Night}}} & \rotatebox{90}{\raisebox{0.075cm}{\textbf{Overcast}}} & \rotatebox{90}{\raisebox{0.075cm}{\textbf{Overcast2}}} & \rotatebox{90}{\raisebox{0.075cm}{\textbf{Rain}}}
& \rotatebox{90}{\raisebox{0.075cm}{\textbf{Dusk}}} & \rotatebox{90}{\raisebox{0.075cm}{\textbf{Night}}} & \rotatebox{90}{\raisebox{0.075cm}{\textbf{Overcast}}} & \rotatebox{90}{\raisebox{0.075cm}{\textbf{Overcast2}}} & \rotatebox{90}{\raisebox{0.075cm}{\textbf{Rain}}}
& \rotatebox{90}{\raisebox{0.075cm}{\textbf{Dusk}}} & \rotatebox{90}{\raisebox{0.075cm}{\textbf{Night}}} & \rotatebox{90}{\raisebox{0.075cm}{\textbf{Overcast}}} & \rotatebox{90}{\raisebox{0.075cm}{\textbf{Overcast2}}} & \rotatebox{90}{\raisebox{0.075cm}{\textbf{Rain}}}
& \rotatebox{90}{\raisebox{0.075cm}{\textbf{Dusk}}} & \rotatebox{90}{\raisebox{0.075cm}{\textbf{Night}}} & \rotatebox{90}{\raisebox{0.075cm}{\textbf{Overcast}}} & \rotatebox{90}{\raisebox{0.075cm}{\textbf{Overcast2}}} & \rotatebox{90}{\raisebox{0.075cm}{\textbf{Rain}}}\\
\hline
 \textbf{References}
& \multicolumn{5}{c|}{\textbf{CosPlace (512D)}}
& \multicolumn{5}{c|}{\textbf{EigenPlaces (512D)}}
& \multicolumn{5}{c|}{\textbf{MixVPR (4096D)}}
& \multicolumn{5}{c|}{\textbf{NetVLAD (4096D)}}
& \multicolumn{5}{c|}{\textbf{SALAD (8448D)}}
& \multicolumn{5}{c}{\textbf{MegaLoc (8448D)}}\\
\hline
 Sun
& 64.7 & 22.8 & 93.1 & 95.4 & 95.5
& 63.0 & 21.2 & 94.5 & 96.2 & 97.1
& 86.6 & 68.2 & 96.5 & \underline{98.0} & 97.4
& 38.4 & 15.2 & 82.2 & 89.9 & 86.6
& 93.7 & 91.3 & 96.8 & \underline{98.2} & 97.7
& 88.6 & 77.8 & 92.6 & 95.5 & 95.5 \\
 Dusk
& - & \underline{34.0} & 66.0 & 66.4 & 68.1
& - & \underline{32.7} & 64.0 & 65.7 & 66.5
& - & \underline{77.6} & 80.5 & 82.1 & 83.4
& - & \underline{29.6} & 39.1 & 35.8 & 37.0
& - & \underline{93.2} & 90.6 & 90.2 & 91.4
& - & \underline{86.4} & 76.2 & 78.3 & 80.1 \\
 Night
& \underline{68.5} & - & 48.3 & 47.7 & 46.6
& 66.2 & - & 43.7 & 43.2 & 39.2
& 85.6 & - & 72.7 & 71.3 & 69.2
& 41.8 & - & 22.4 & 19.7 & 18.1
& 92.4 & - & 89.0 & 87.8 & 88.7
& 86.9 & - & 65.1 & 68.1 & 69.4 \\
Overcast
& 67.8 & 30.7 & - & \underline{96.9} & \underline{96.1}
& \underline{67.7} & 30.0 & - & \underline{97.4} & \underline{97.3}
& \underline{89.7} & 76.6 & - & 97.9 & \underline{98.2}
& \underline{47.7} & 20.9 & - & \underline{91.8} & 87.6
& 94.5 & 91.2 & - & \underline{98.2} & \underline{98.1}
& 87.9 & 77.5 & - & \underline{97.4} & \underline{96.9} \\
 Overcast2
& 66.8 & 27.4 & \underline{96.6} & - & 95.8
& 64.8 & 23.0 & \underline{96.9} & - & 96.4
& 86.0 & 71.5 & 97.5 & - & 98.0
& 41.5 & 17.8 & \underline{89.3} & - & \underline{88.3}
& \underline{94.8} & 91.8 & 97.7 & - & 98.0
& 88.8 & 79.3 & \underline{95.5} & - & 95.9 \\
 Rain
& 65.0 & 25.0 & 95.2 & 95.4 & -
& 64.3 & 25.1 & 96.6 & 96.5 & -
& 85.5 & 63.6 & \underline{97.9} & 97.8 & -
& 44.1 & 14.8 & 84.7 & 88.3 & -
& 94.0 & 91.0 & \underline{97.8} & 98.0 & -
& \underline{91.6} & 78.8 & 95.3 & 96.6 & - \\

\hline

\multicolumn{31}{l}{\textbf{\emph{Score-level baselines}}} \\

dMat. Avg.~\cite{fischer2020event}
& 74.8 & 36.0 & 94.4 & 96.4 & 95.5
& 75.1 & 37.2 & 95.7 & 97.0 & 96.4
& 95.0 & 87.1 & 98.7 & 98.7 & 98.6
& 67.2 & 35.4 & 92.5 & 96.1 & 94.2
& 98.1 & 96.2 & 98.3 & 98.8 & 98.8
& 95.6 & 89.7 & 95.5 & 97.5 & 97.4 \\
dMat. Min.~\cite{fischer2020event}
& 77.5 & 34.0 & \textbf{98.0} & 97.9 & 98.4
& 74.3 & 32.7 & \textbf{98.3} & \textbf{98.5} & 98.4
& 93.0 & 78.1 & 98.6 & 98.6 & 98.9
& 54.3 & 30.1 & 92.2 & 94.7 & 91.3
& 96.5 & 94.2 & 98.8 & 98.7 & 98.9
& 94.0 & 87.3 & 97.1 & 98.0 & 98.0 \\
 Std. dMat. Min.
& 72.3 & 31.0 & 96.3 & 97.0 & 97.9
& 70.9 & 28.2 & 97.7 & 98.2 & \textbf{98.5}
& 90.7 & 80.2 & 98.7 & 98.5 & 98.8
& 57.3 & 28.8 & 90.0 & 94.2 & 92.5
& 96.9 & 94.9 & 98.7 & 98.7 & 98.8
& 93.3 & 87.3 & 96.1 & 98.0 & 97.6 \\
 dMat. Med.~\cite{fischer2020event}
& 70.1 & 29.6 & 93.4 & 95.6 & 94.8
& 69.2 & 28.8 & 94.7 & 96.3 & 95.8
& 91.8 & 78.8 & 96.9 & 97.6 & 97.7
& 57.0 & 25.1 & 85.8 & 91.3 & 89.0
& 97.3 & 94.2 & 97.3 & 98.3 & 97.9
& 92.6 & 84.4 & 93.4 & 95.7 & 95.7 \\
 BSF~\cite{molloy2021intelligent}
& 77.7 & 34.0 & 97.6 & 97.8 & 98.2
& 74.7 & 32.6 & 98.2 & 98.3 & 98.2
& 93.5 & 78.7 & 98.7 & 98.6 & 98.9
& 58.4 & 31.2 & 92.4 & 95.0 & 93.3
& 96.8 & 94.9 & 98.6 & 98.6 & 98.8
& 94.3 & 88.1 & 96.7 & 97.8 & 98.0 \\

\hline
\multicolumn{31}{l}{\textbf{\emph{Descriptor-level baselines}}} \\

Pooling
& 77.5 & 34.0 & \textbf{98.0} & 97.9 & 98.4
& 74.3 & 32.7 & \textbf{98.3} & \textbf{98.5} & 98.4
& 93.0 & 78.1 & 98.6 & 98.6 & 98.9
& 54.3 & 30.1 & 92.2 & 94.7 & 91.3
& 96.5 & 94.2 & 98.8 & 98.7 & 98.9
& 94.0 & 87.3 & 97.1 & 98.0 & 98.0 \\

HOPS~\cite{malone2025hyperdimensional}
& 74.7 & 30.8 & 96.5 & 97.8 & 98.1
& 73.0 & 30.5 & 97.4 & 98.2 & 98.0
& 94.0 & 83.8 & \textbf{98.9} & 99.0 & 98.9
& 67.2 & 39.1 & 94.7 & 97.2 & 96.2
& 98.1 & 96.3 & 98.6 & 98.9 & 99.0
& 95.9 & 89.9 & 97.7 & 98.5 & 98.6 \\

DisPlace (Ours)
& \textbf{85.7} & \textbf{60.6} & 97.6 & \textbf{98.3} & \textbf{98.7}
& \textbf{82.6} & \textbf{54.0} & 97.9 & 98.4 & 98.4
& \textbf{97.4} & \textbf{92.1} & \textbf{98.9} & \textbf{99.1} & \textbf{99.0}
& \textbf{81.3} & \textbf{55.4} & \textbf{96.8} & \textbf{98.5} & \textbf{97.9}
& \textbf{98.5} & \textbf{97.3} & \textbf{98.9} & \textbf{99.0} & \textbf{99.2}
& \textbf{97.3} & \textbf{93.1} & \textbf{98.4} & \textbf{98.7} & \textbf{99.1} \\

\bottomrule
\end{tabular}
\label{tab:robotcar_table}
\vspace*{-0.15cm}
\end{table*}
}

\textit{Score-level methods} first compute query-reference distances independently for each traversal before aggregating the resulting scores.
dMat.~Avg.~\cite{fischer2020event}, dMat.~Min.~\cite{fischer2020event}, and dMat.~Med.~\cite{fischer2020event} aggregate distances using averaging, minimum, and median operations, respectively.
We additionally introduce Std.~dMat.~Min., which standardizes distance scores within each traversal prior to minimum aggregation to reduce sensitivity to traversal-specific score distributions.
Finally, we replicate Bayesian Selective Fusion (BSF)~\cite{molloy2021intelligent}, which selects informative traversals based on query similarity before computing a fused posterior estimate.

\subsection{Evaluation Metrics}
\label{subsec:metrics}

We evaluate retrieval performance using Recall@1, which measures the proportion of queries for which the top-ranked retrieved place is correct.
Ground-truth tolerances follow standard evaluation protocols where available: one-to-one correspondence for Nordland~\cite{somayeh2025improving, malone2025hyperdimensional}, a 25\,m UTM radius for Pittsburgh30k~\cite{hausler2021patchnetvlad, berton2025megaloc, berton2023eigenplaces}, and exact landmark identity for GLDv2-micro~\cite{malone2025hyperdimensional}.
For Oxford RobotCar, we use a tolerance of $\pm5$ frames ($\approx$5\,m).
\section{Results}
\label{sec:results}

We evaluate DisPlace across appearance-varying, viewpoint-varying, and unstructured multi-reference VPR settings.
We then analyze the effect of dimensionality reduction, the variance threshold $\tau$, qualitative retrieval behavior, interpretability, and inference-time efficiency.

\subsection{Appearance-Varying Datasets}
\label{subsec:results_appearance}

The following experiment evaluates DisPlace under severe appearance variation to assess whether the proposed discriminative projection improves multi-reference retrieval robustness. \Cref{tab:robotcar_table,tab:nordland_table} report Recall@1 on Oxford RobotCar and Nordland, respectively, comparing DisPlace against single-reference and multi-reference baselines under appearance variation.

Across both datasets, multi-reference methods generally outperform the best single-reference result; specifically, DisPlace exceeds the best single-reference result in all 54 appearance-varying conditions.
Among multi-reference methods, DisPlace achieves the best performance in 49 out of 54 cases, including 25 out of 30 on RobotCar and all 24 on Nordland.

DisPlace consistently improves over HOPS~\cite{malone2025hyperdimensional}, the current descriptor-level fusion state-of-the-art baseline, across descriptors and datasets.
The largest gains occur under the most challenging conditions, namely Night on RobotCar and Winter on Nordland, where appearance and illumination changes are most severe.

Despite storing only a single fused descriptor per place, DisPlace also outperforms Pooling and the score-level baselines in the majority of cases, although these methods retain all $K$ reference descriptors per place during inference. In the 4 cases where DisPlace trails Pooling and dMat.~Min., it trails by only $\approx$0.25 percentage points on average, while requiring substantially lower storage and inference time (see~\Cref{subsec:complexity}).
Pooling and dMat.~Min.\ produce identical Recall@1 in all experiments because both select the globally closest reference image to the query; both are reported for consistency with prior multi-reference VPR evaluations~\cite{fischer2020event, malone2025hyperdimensional}.

\begin{table}[t]
\caption{Recall@1 results on Pittsburgh30k (top) and GLDv2-micro (bottom). Best results are \textbf{bolded}, and the second-best results are \underline{underlined}.}
\vspace*{-0.1cm}
\centering
\scriptsize
\setlength{\tabcolsep}{1.9pt}
\renewcommand{\arraystretch}{1.0}

\begin{tabular}{c l cccccc}
\toprule
& \textbf{Method} 
& \rotatebox{90}{\textbf{CosPlace}} 
& \rotatebox{90}{\textbf{EigenPlaces}} 
& \rotatebox{90}{\textbf{MixVPR}} 
& \rotatebox{90}{\textbf{NetVLAD}} 
& \rotatebox{90}{\textbf{SALAD}} 
& \rotatebox{90}{\textbf{MegaLoc}} \\
\midrule

\multicolumn{1}{c}{} & \multicolumn{7}{l}{\textbf{\emph{Score-level baselines}}} \\
\multirow{8}{*}{\rotatebox{90}{\textbf{Pittsburgh30k}}}
& dMat. Avg.~\cite{fischer2020event}       & 58.9 & 65.6 & 72.2 & 63.4 & 74.3 & 80.7 \\
& dMat. Min.~\cite{fischer2020event}       & \underline{90.2} & \underline{91.9} & \textbf{91.6} & \textbf{84.9} & \underline{92.4} & \textbf{93.9} \\
& Std. dMat. Min.  & 74.1 & 83.2 & 84.9 & 82.2 & 83.8 & 87.4 \\
& dMat. Med.~\cite{fischer2020event}       & 22.7 & 31.1 & 28.3 & 31.2 & 33.7 & 38.5 \\
& BSF~\cite{molloy2021intelligent}         & \textbf{90.3} & \textbf{92.0} & \underline{91.3} & \underline{83.4} & \textbf{92.7} & \underline{93.9} \\

\cmidrule(lr){2-8}
\multicolumn{1}{c}{} & \multicolumn{7}{l}{\textbf{\emph{Descriptor-level baselines}}} \\

& Pooling          & \underline{90.2} & \underline{91.9} & \textbf{91.6} & \textbf{84.9} & \underline{92.4} & \textbf{93.9} \\

& HOPS~\cite{malone2025hyperdimensional}  & 67.7 & 72.4 & 77.4 & 67.3 & 80.1 & 84.9 \\

& DisPlace (Ours)         & 81.1 & 82.5 & 85.3 & 66.8 & 91.4 & 91.5 \\

\midrule

\multicolumn{1}{c}{} & \multicolumn{7}{l}{\textbf{\emph{Descriptor-level baselines}}} \\
\multirow{3}{*}{\rotatebox{90}{\shortstack[c]{\textbf{GLDv2}\\\textbf{micro}}}}
&  Pooling  & \underline{59.2} & \textbf{63.5} & \underline{52.2} & \underline{57.2} & \underline{69.7} & \textbf{85.6} \\

& HOPS~\cite{malone2025hyperdimensional}     & 51.2 & 55.3 & 48.3 & 55.2 & 65.8 & 83.2 \\

& DisPlace (Ours) & \textbf{59.3} & \underline{60.2} & \textbf{54.0} & \textbf{61.0} & \textbf{75.3} & \underline{84.9} \\

\bottomrule
\end{tabular}

\label{tab:pitts30k_gldv2_table}
\vspace*{-0.2cm}
\end{table}

\subsection{Viewpoint-Varying Dataset}
\label{subsec:results_viewpoint}

The next experiment evaluates DisPlace under extreme viewpoint variation, where reference observations of the same place may share little or no visual overlap. \Cref{tab:pitts30k_gldv2_table} reports Recall@1 on Pittsburgh30k, where the $K = 24$ reference sets correspond to distinct viewpoints rather than appearance conditions (2 pitch angles and 12 yaw angles). 
Pittsburgh30k is particularly challenging because opposing viewpoints may share little visual overlap.

Pooling and score-level methods achieve the highest absolute Recall@1 because they retain all individual viewpoint descriptors and can select the closest matching view at query time.
By contrast, descriptor-level fusion methods must aggregate all 24 viewpoints into a single per-place representation, including opposing views with little or no visual overlap.
This makes descriptor-level fusion substantially more challenging on Pittsburgh30k than on the appearance-varying benchmarks, where all traversals observe broadly similar scene geometry.

Within descriptor-level fusion methods, DisPlace outperforms HOPS for five out of six descriptors, with relative improvements of approximately 8--20\% for CosPlace, EigenPlaces, MixVPR, SALAD, and MegaLoc, while trailing HOPS marginally when NetVLAD is used as the underlying descriptor.

These results suggest that the derived projection retains descriptor dimensions that are informative for place identity while suppressing dimensions dominated by viewpoint variation, a distinction that descriptor aggregation in the original feature space cannot explicitly make.

\subsection{Unstructured Multi-Reference Dataset}
\label{subsec:results_unstructured}

\Cref{tab:pitts30k_gldv2_table} also reports Recall@1 on GLDv2-micro, where each landmark has 7--9 reference images captured under irregular viewpoint and appearance variation.
Because the dataset lacks structured per-reference-set correspondence, we follow~\cite{malone2025hyperdimensional} and evaluate descriptor-level methods only: Pooling, HOPS, and DisPlace.

DisPlace outperforms HOPS across all descriptors, with absolute Recall@1 improvements of up to 10\% on average.
It also outperforms Pooling on four out of six descriptors, trailing by $\approx$3\% on average in the remaining two cases, despite only storing a single fused descriptor per place.

\subsection{Analysis: Effect of Dimensionality Reduction}
\label{subsec:analysis_dim_red}

\begin{figure}[t]
    \centering
    \includegraphics[width=0.99\linewidth]{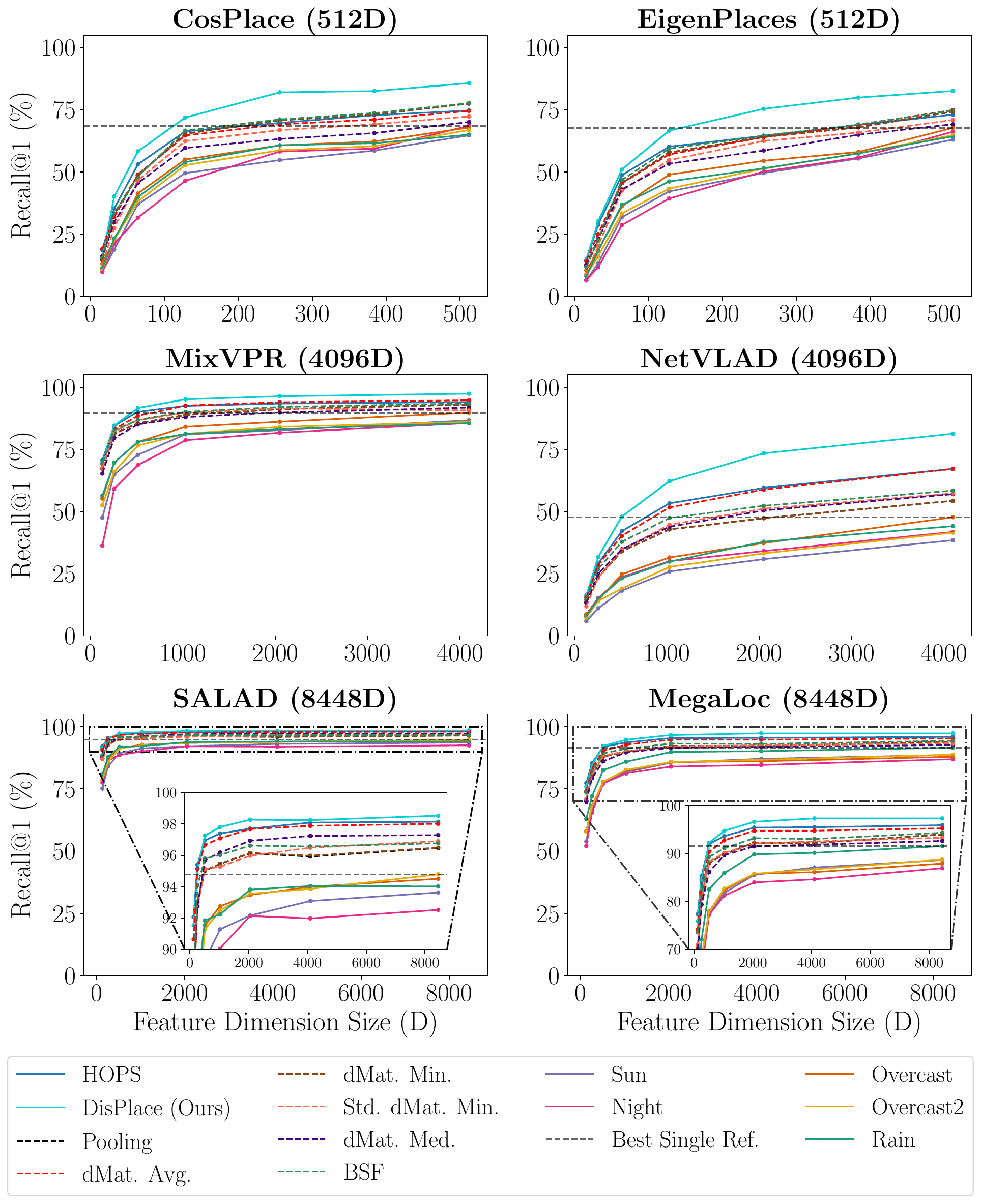}
    \vspace*{-0.4cm}
    \caption{\textbf{Effect of dimensionality reduction on Recall@1 for the Oxford RobotCar Dusk query across six VPR descriptors.}
    Each subplot shows Recall@1 as a function of descriptor dimension after Gaussian Random Projection.
    DisPlace achieves the highest full-dimensional Recall@1 across all descriptors and matches or exceeds the best full-dimensional single- and multi-reference baselines even after substantial descriptor compression.}
    \label{fig:analysis_dim_red}
    \vspace*{-0.15cm}
\end{figure}

We now evaluate whether the discriminative projection remains effective under substantial descriptor compression. Descriptor storage and retrieval cost scale with dimensionality, making post-hoc compression important for large-scale VPR deployment~\cite{berton2022rethinking, berton2022benchmark}.
Following~\cite{malone2025hyperdimensional}, we apply Gaussian Random Projection to all methods to reduce dimensionality.

\Cref{fig:analysis_dim_red} reports Recall@1 on the Oxford RobotCar Dusk query across all six descriptors.
DisPlace matches or exceeds the best full-dimensional single- and multi-reference baselines across the tested descriptor dimensionalities. 
This indicates that the discriminative projection in DisPlace identifies compact subspaces that preserve the information most relevant for multi-reference VPR.

\subsection{Ablation: Variance Threshold \texorpdfstring{$\tau$}{tau}}
\label{subsec:ablation}

\begin{figure}[t]
    \centering
    \includegraphics[width=0.99\linewidth]{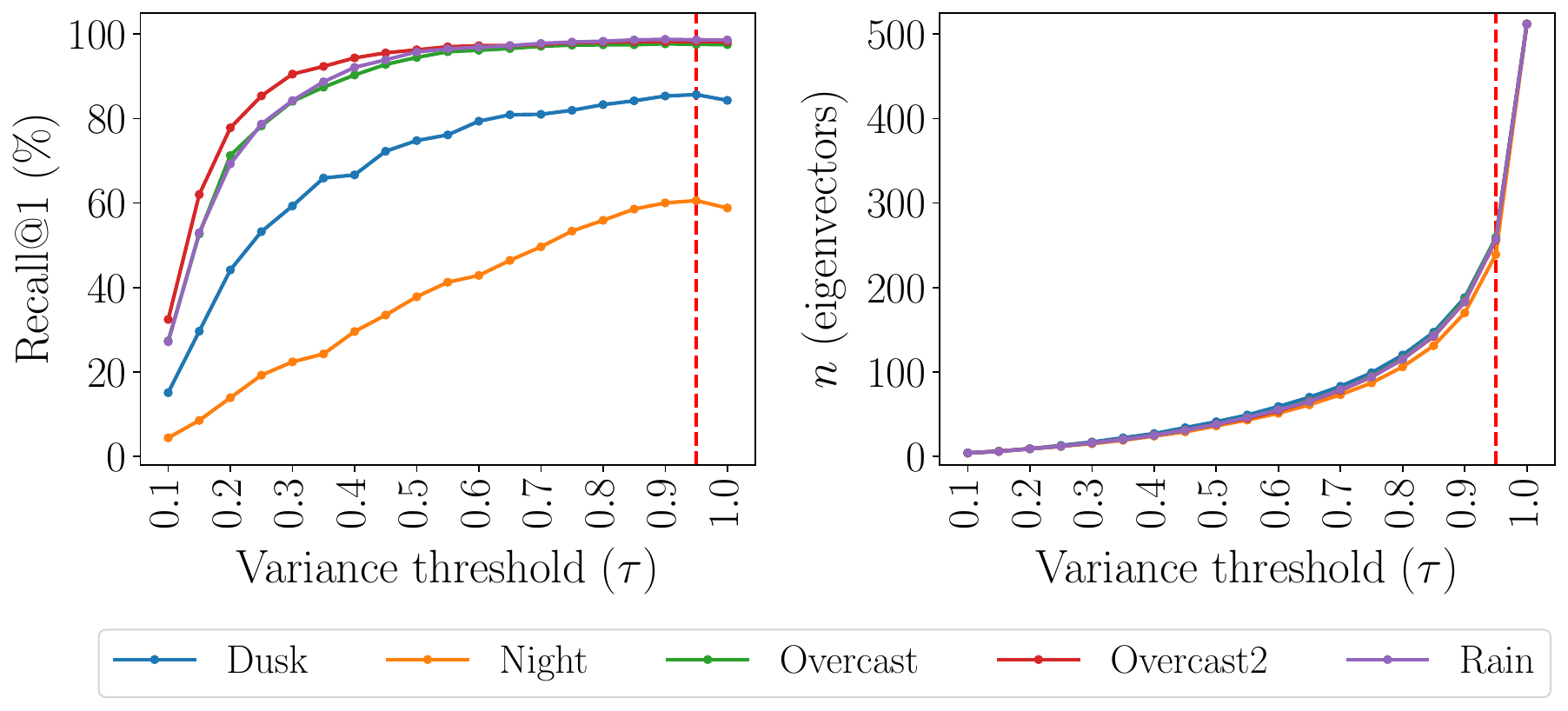}
    \vspace*{-0.4cm}
    \caption{\textbf{Ablation of the variance threshold $\tau$ on Oxford RobotCar using CosPlace ($D = 512$).}
    \emph{Left:} Recall@1 as a function of $\tau$ for each query condition.
    \emph{Right:} Projected dimension $n$ as a function of $\tau$.
    Recall@1 rapidly saturates and remains stable across a broad range of $\tau$, indicating that most discriminative structure is captured by a relatively small subset of projection directions.
    By contrast, the projected dimensionality increases sharply as $\tau \rightarrow 1$, showing that the remaining directions contribute little retrieval benefit despite substantially increasing descriptor size.
    At $\tau = 0.95$, DisPlace retains approximately 50\% of the original descriptor dimensionality without performance degradation.}
    \label{fig:ablation}
    \vspace*{-0.2cm}
\end{figure}

The variance threshold $\tau$ controls the number of projection directions retained in $\mathbf{P}$ (\Cref{eq:n_select}).
\Cref{fig:ablation} reports Recall@1 and projected dimension $n$ as a function of $\tau$ on Oxford RobotCar using CosPlace ($D=512$).

Recall@1 increases rapidly up to approximately $\tau = 0.8$ and then plateaus across all query conditions, indicating that most discriminative structure is concentrated within a relatively small subset of projection directions.
Beyond this point, increasing $\tau$ primarily retains low-contribution directions that substantially increase descriptor dimensionality while providing little or no retrieval benefit.
At the chosen value $\tau = 0.95$, DisPlace retains $n = 239$ to $259$ dimensions across the five query conditions, corresponding to an approximately 50\% reduction from the original $D = 512$ descriptor size, while matching or slightly exceeding performance at $\tau = 1.0$.
Retaining all positive-eigenvalue directions ($\tau = 1.0$) provides no performance benefit and can even slightly degrade retrieval, indicating that the final directions primarily capture within-place variation rather than stable place-discriminative information.

\subsection{Qualitative Results}
\label{subsec:qual_results}

\begin{figure}[t]
    \centering
    \includegraphics[width=\columnwidth]{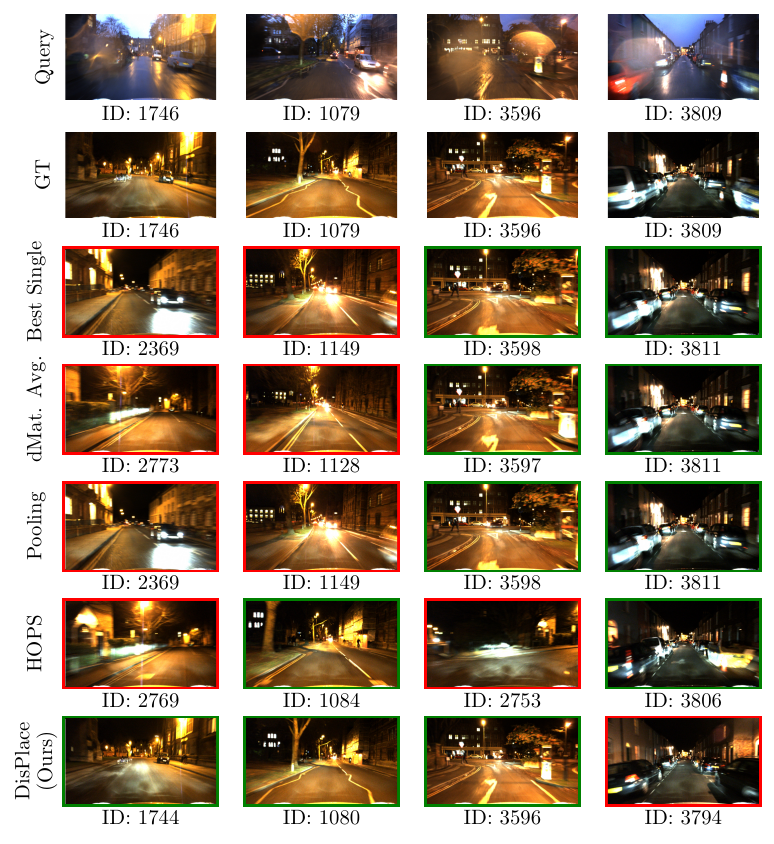}
    \vspace*{-0.5cm}
    \caption{\textbf{Qualitative retrieval results on Oxford RobotCar Dusk using CosPlace.}
    Each column shows a query image (\emph{top}), its ground-truth match (\emph{second row}), and the top-1 retrieval from representative single-reference and multi-reference methods.
    {\color{forestgreen} {\textbf{Green}}} borders indicate correct matches (within $\pm5$ frames); {\color{red} {\textbf{red}}} borders indicate incorrect matches.
    The examples illustrate cases where DisPlace uniquely succeeds (\textit{column 1}), where descriptor-level fusion outperforms score-level methods (\textit{column 2}), where DisPlace and score-level methods succeed while HOPS fails (\textit{column 3}; note that DisPlace requires only a fraction of the storage and compute of score-level methods), and one representative failure case (\textit{column 4}).}
    \label{fig:qual_results}
    \vspace*{-0.2cm}
\end{figure}

\Cref{fig:qual_results} shows representative Oxford RobotCar Dusk retrieval examples using CosPlace.
The examples compare the best single-reference baseline, dMat.~Avg., Pooling, HOPS, and DisPlace.
They illustrate cases where DisPlace uniquely retrieves the correct place, where descriptor-level fusion succeeds while score-level methods fail, where DisPlace and score-level methods succeed despite HOPS failing, and one representative failure case.

\subsection{Analysis: Interpretability}

\begin{figure}
    \centering

    \includegraphics[width=0.99\linewidth]{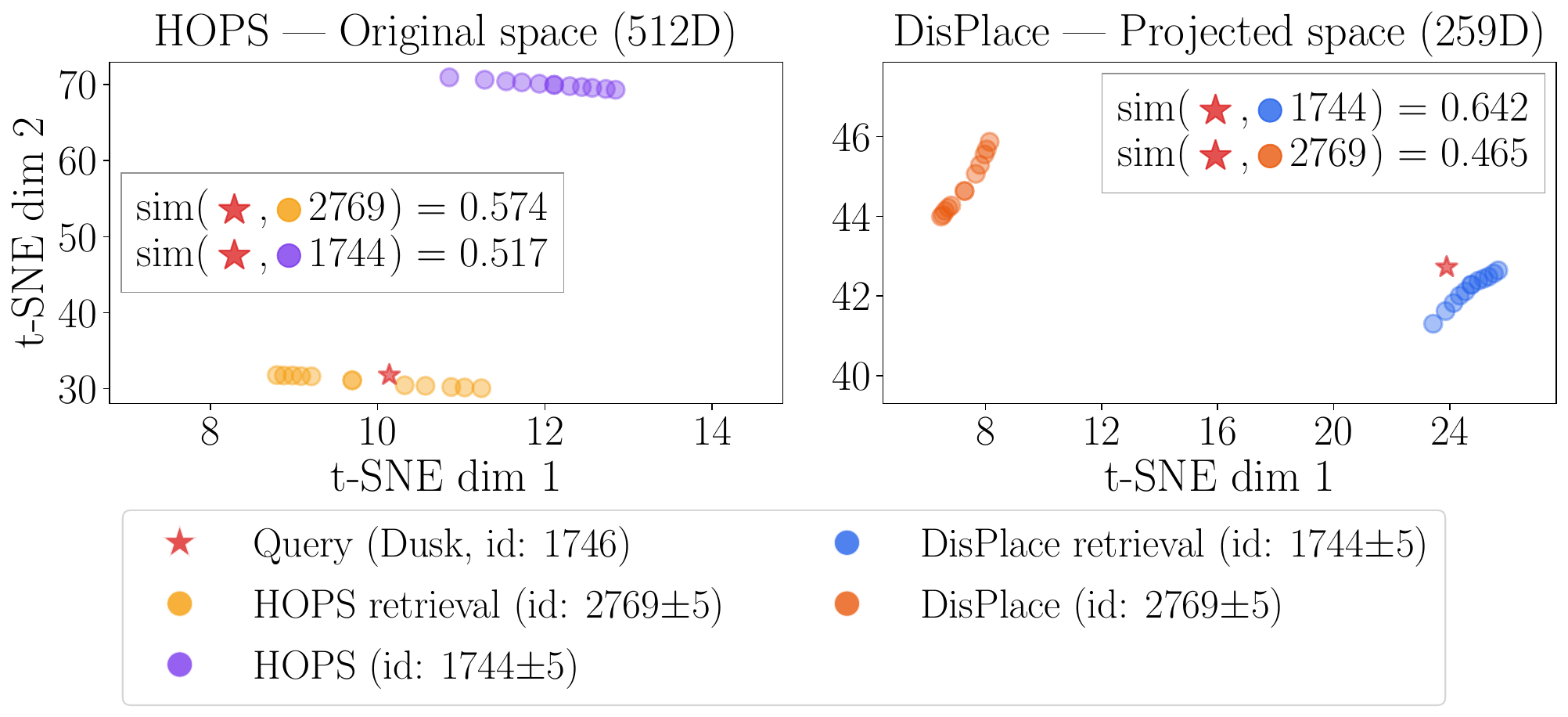}
    \vspace*{-0.25cm}
    \caption{\textbf{t-SNE visualization of descriptor spaces for query 1746 (Dusk) on Oxford RobotCar using CosPlace.}
    \emph{Left:} Original descriptor space (512-D), where HOPS incorrectly retrieves id 2769 (\textcolor{orange}{orange}), which lies closer to the query (\textcolor{red}{red star}) than the correct place (\textcolor{violet}{purple}, id: $1744 \pm 5$).
    \emph{Right:} DisPlace projected space (259-D). The correct place (\textcolor{blue}{blue}, id: $1744 \pm 5$) becomes nearest to the query while the \textcolor{orange}{confounding place} is pushed away.}

    \label{fig:mech_analysis}
\end{figure}

To illustrate the effect of the derived projection, \Cref{fig:mech_analysis} visualizes the original and projected descriptor spaces using t-SNE for query 1746 under the Oxford RobotCar Dusk condition (see \Cref{fig:qual_results}, Column 1) with CosPlace. In the original descriptor space, the incorrect HOPS retrieval lies closer to the query than the correct place.
After projection into the DisPlace discriminative space, the correct place becomes nearest to the query, while the previously confounding place is pushed farther away.
This illustrates how DisPlace increases similarity to stable same-place representations while suppressing visually confounding places, a structural change that element-wise summation in the original space (as in HOPS) cannot achieve.

\subsection{Computational Complexity during Inference}
\label{subsec:complexity}

\begin{figure}[t]
    \centering
    \includegraphics[width=0.99\linewidth]{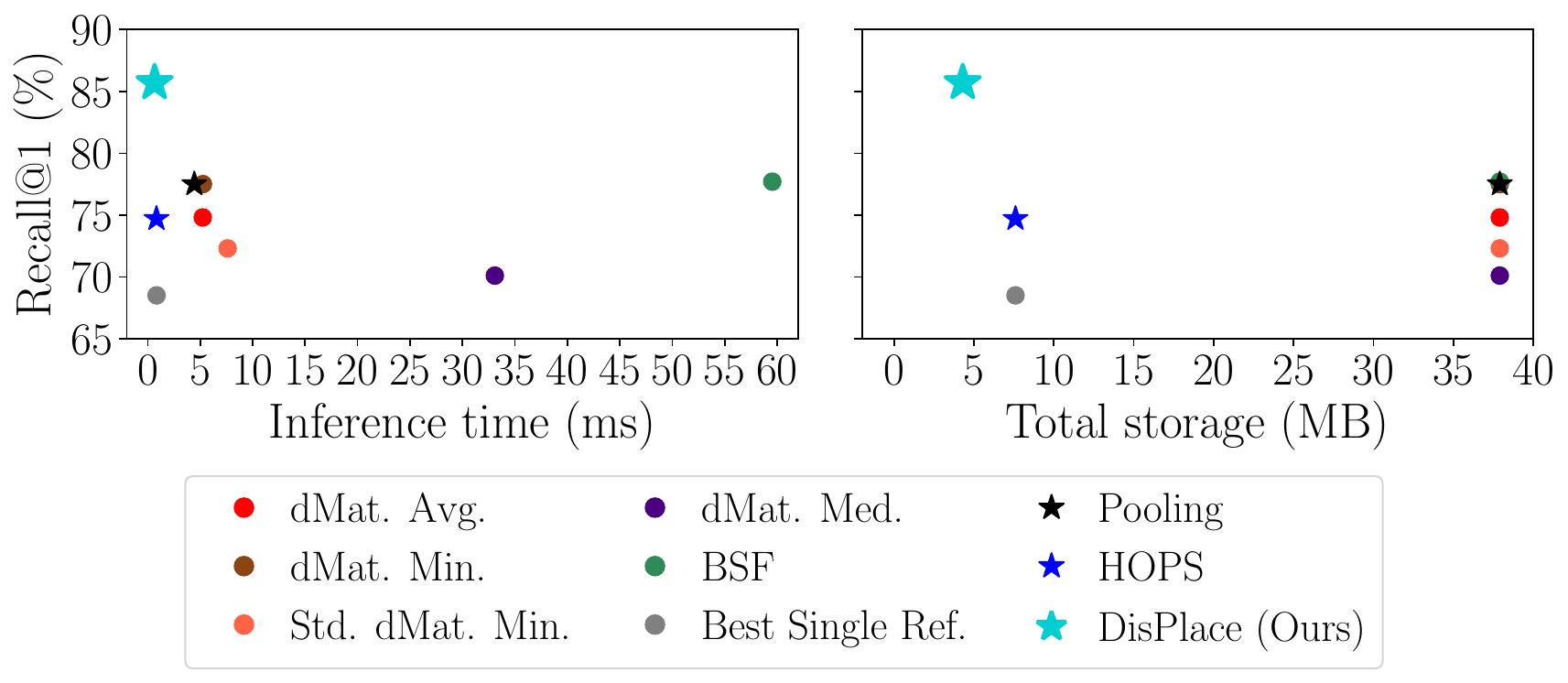}
    \vspace*{-0.25cm}
    \caption{\textbf{Recall@1 versus inference time (\emph{left}) and total storage (\emph{right}) on Oxford RobotCar Dusk using CosPlace.}
    DisPlace achieves the highest Recall@1 while requiring 0.65\,ms inference time and 4.3\,MB storage.
    Score-level methods require 5--60\,ms inference time and 37.9\,MB storage.}

    \label{fig:complexity_results}

\end{figure}

All methods are implemented in PyTorch and evaluated on a single NVIDIA GeForce RTX 4080 GPU; timings are averaged over 100 runs.
\Cref{fig:complexity_results} reports Recall@1 versus inference time and storage on Oxford RobotCar using CosPlace ($D = 512$, $K = 5$, $N = 3{,}876$) for the Dusk query.

Score-level methods and pooling retain all $K$ reference descriptors per place, requiring $O(K\cdot N\cdot D)$ storage. 
In contrast, DisPlace stores a single $n$-dimensional descriptor per place, together with a fixed projection matrix $\mathbf{P} \in \mathbb{R}^{D \times n}$.
For $\tau = 0.95$, DisPlace uses $n = 259$, requiring 4.3\,MB storage and 0.65\,ms inference time, compared with 7.6\,MB and 0.83\,ms for HOPS and 37.9\,MB with 4.4--60\,ms for Pooling and score-level methods.

DisPlace reduces inference time and storage by 22\% and 43\%, respectively, relative to HOPS, while improving Recall@1 by 11 percentage points.
Compared with Pooling and score-level methods, DisPlace reduces inference time by 85--99\% and storage by 89\%, while improving Recall@1 by 8--16 percentage points.
\section{Discussion and Conclusion}
\label{sec:discuss_and_conclude}

In this paper, we presented DisPlace, a discriminative projection framework for multi-reference VPR that formulates descriptor fusion as a generalized eigenvalue problem over within- and between-place scatter matrices. DisPlace operates post-hoc on pretrained VPR descriptors without retraining.

Unlike descriptor aggregation approaches such as HOPS~\cite{malone2025hyperdimensional}, which fuse reference descriptors directly in the original feature space, DisPlace first derives a discriminative projection from variation across places and reference traversals. This projection identifies directions that preserve place-discriminative information while suppressing condition- and viewpoint-induced variation prior to aggregation.

Through evaluation on four datasets across six VPR descriptors, we demonstrated that DisPlace outperforms existing multi-reference baselines in 49 out of 54 appearance-varying conditions, consistently improves descriptor-level fusion under viewpoint and unstructured settings, and achieves lower storage and inference complexity than all compared methods while maintaining strong retrieval performance.

The results further reveal that the discriminative projection is most effective when the variation observed across the reference traversals is representative of the variation encountered at query time, particularly under appearance variation where descriptor changes exhibit structured and suppressible patterns.
Under extreme viewpoint variation, where opposing views may share little or no visual overlap, DisPlace still improves over HOPS in five out of six descriptors but trails score-level methods that retain all individual reference descriptors.
This suggests that discriminative projection is most beneficial when within-place variation can be separated from stable place information within a shared descriptor space, rather than when observations correspond to fundamentally different visual content or previously unseen variations.
Bridging the remaining gap between compact descriptor-level fusion and score-level methods under severe viewpoint variation remains an important direction for future work.

Several extensions of DisPlace are possible.
A natural direction is extending the framework from single-image retrieval to sequence-to-sequence matching, where the derived projection could be integrated within multi-sequence localization frameworks such as~\cite{Vysotska2019effective, garg2022seqmatchnet}.
Another promising direction is viewpoint-aware fusion, where separate projections are derived for subsets of reference viewpoints with sufficient visual overlap, enabling discriminative fusion within geometrically consistent viewpoint groups rather than across all viewpoints indiscriminately.

{\small
\bibliographystyle{IEEEtran}
\bibliography{references}
}

\end{document}